\documentclass[nonacm,sigconf]{acmart}

\usepackage{booktabs}
\usepackage{multirow}
\usepackage{makecell}
\usepackage{subcaption}
\usepackage{threeparttable} 
\usepackage{tabularx}
\usepackage{textcomp}
\AtBeginDocument{%
  }

\usepackage{caption}

\usepackage{xurl} 

\usepackage{xspace} 
\usepackage{listings, mdframed}
\newcommand{\name}{ImpReSS\xspace}
\usepackage{enumitem} 
\usepackage{graphicx} 
\usepackage{dblfloatfix} 

\begin{document}

\title{\textbf{\name: Implicit Recommender System for Support Conversations}}

\author{Omri Haller}
\affiliation{%
  \institution{Ben Gurion University of The Negev}
  \city{Be'er Sheva}
  \country{Israel}}
\email{haller@post.bgu.ac.il}

\author{Yair Meidan}
\affiliation{%
  \institution{Ben Gurion University of The Negev}
  \city{Be'er Sheva}
  \country{Israel}}
\email{yair.meidan@gmail.com}

\author{Dudu Mimran}
\affiliation{%
  \institution{Ben Gurion University of The Negev}
  \city{Be'er Sheva}
  \country{Israel}}
\email{dudu@dudumimran.com}

\author{Yuval Elovici}
\affiliation{%
  \institution{Ben Gurion University of The Negev}
  \city{Be'er Sheva}
  \country{Israel}}
\email{elovici@bgu.ac.il}

\author{Asaf Shabtai}
\affiliation{%
  \institution{Ben Gurion University of The Negev}
  \city{Be'er Sheva}
  \country{Israel}}
\email{shabtaia@bgu.ac.il}

\begin{abstract}
Following recent advancements in large language models (LLMs), LLM-based chatbots have transformed customer support by automating interactions and providing consistent, scalable service. 
While LLM-based conversational recommender systems (CRSs) have attracted attention for their ability to enhance the quality of recommendations, limited research has addressed the \emph{implicit} integration of recommendations within customer support interactions.
In this work, we introduce \name, an implicit recommender system designed for customer support conversations. 
\name operates alongside existing support chatbots, where users report issues and chatbots provide solutions. 
Based on a customer support conversation, \name identifies opportunities to recommend relevant solution product categories (SPCs) that help resolve the issue or prevent its recurrence -- thereby also supporting business growth.
Unlike traditional CRSs, \name functions entirely implicitly and does not rely on any assumption of a user's purchasing intent.
Our empirical evaluation of \name's ability to recommend relevant SPCs that can help address issues raised in support conversations shows promising results, including an MRR@1 (and recall@3) of 0.72 (0.89) for general problem solving, 0.82 (0.83) for information security support, and 0.85 (0.67) for cybersecurity troubleshooting. 
To support future research, our data and code will be shared upon request.
\end{abstract}

\maketitle 

\section{Introduction}\label{sec:Introduction}
Large language models (LLMs) have become transformative tools in numerous domains, fundamentally changing the way we approach complex tasks in almost all areas of life~\cite{zhao2025survey}.
Among their countless applications, the integration of LLMs in customer support has been particularly impactful~\cite{shareef2024enhancing}, allowing businesses to provide 24/7 assistance and enhance customer satisfaction by improving the speed, consistency, and accuracy of customer interactions~\cite{dinne2024llms}.

Previous research on LLM-based customer support chatbots has concentrated on improving service quality and reducing human agents' workload with self-service tools and automated responses~\cite{shareef2024enhancing}.
However, the opportunity to leverage customer interactions for cross-selling and product recommendations~\cite{dinne2024llms} has largely been overlooked.
As a result, businesses have missed out on value creation opportunities that could benefit them and their clients.
Research on conversational recommender systems (CRSs) has led to advancements in the use of dialogue to understand user preferences and suggest appropriate items~\cite{lin2025can}.
However, these systems face challenges in customer support contexts, where interactions are primarily problem-focused rather than sales-oriented.
In such cases, user preferences are often unavailable, and explicitly requesting them may be inappropriate or irrelevant to the situation, and potentially harm the user experience.

\begin{figure}[!t]
  \centering
  \includegraphics[width=0.9\linewidth, trim={0 0 0 0},clip]{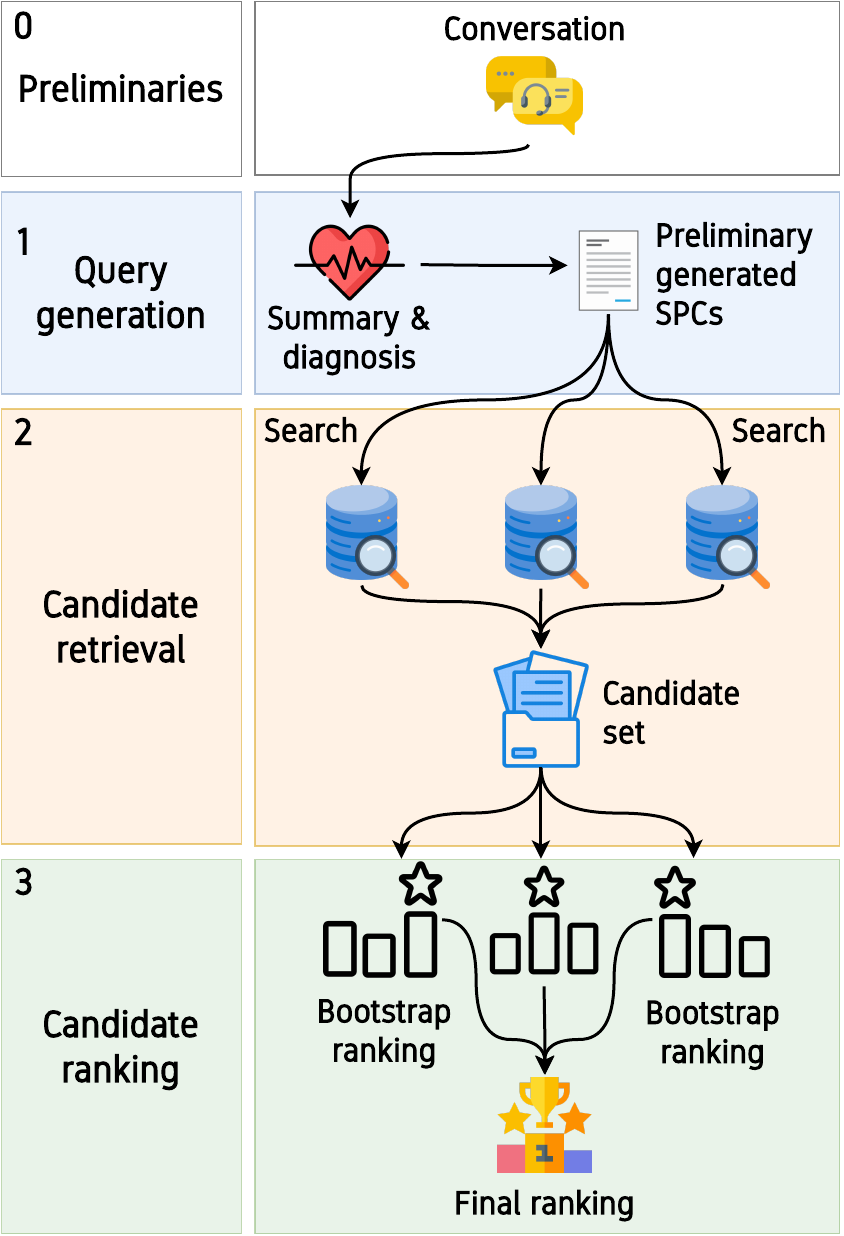}
    \vspace{5pt}
  \setlength{\abovecaptionskip}{0pt}
  \setlength{\belowcaptionskip}{0pt}
  \caption{The proposed method's pipeline\protect\footnotemark.}
  \label{fig:methodology}
\end{figure}

To address this gap, we developed \textbf{\name}: an \textbf{imp}licit \textbf{re}co-
mmender \textbf{s}ystem for \textbf{s}upport conversations.
\name employs a novel approach that augments support-oriented conversations with product (or service) recommendations by implicitly identifying users' \emph{needs} during the problem-solving stage, rather than users' preferences. 
In addition, unlike previous CRS research, \name does not assume any user purchasing intent.
As shown in \autoref{fig:methodology}, \name consists of three main steps: (1) \emph{Query Generation}, in which an LLM generates a brief summary (with a diagnosis) of the conversation, as well as a set of preliminary solution product categories (SPCs) derived from the support conversation's summary and diagnosis; (2) \emph{Candidate Retrieval}, in which the query is used to search designated catalogs for the most relevant SPCs; and (3) \emph{Candidate Ranking}, in which the retrieved SPCs are prioritized.

To integrate \name into real-world business workflows, organizations can map general SPCs to their specific company products (brands), and implement various presentation strategies for the recommendations.
In the (\emph{"In-Chat"}) presentation strategy, the chatbot suggests the top-ranked candidate as a natural continuation of the support conversation, following the resolution of the user's issue.
Alternatively, multiple top-ranked candidates can be displayed below the conversation interface, similar to e-commerce platforms, under a heading such as \textit{"Users who encountered this problem found these products useful"}. 
This \emph{"Related Items"} strategy is applicable to both chatbots and online forums operated by commercial entities.

We evaluate \name using three conversational support datasets from various domains.
Our empirical results highlight \name's promising SPC recommendation capabilities and ability to make accurate recommendations without relying on user preference data. 
\name achieved MRR@1 values (and recall@3) of 0.72 (0.89) for general problem solving, 0.82 (0.83) for information security support, and 0.85 (0.67) for cybersecurity troubleshooting.

The contributions of this research can be summarized as follows:

\begin{enumerate}[nosep, leftmargin=*]
    \item We introduce \name, a novel LLM-based method for implicit SPC recommendations in conversational support settings, which assumes no user purchasing intent and requires no user preferences or background information as input.
    \item We empirically evaluate \name using multiple datasets, and show promising performance, achieved early in a conversation.
    \item To promote future research, our data and code will be shared upon request.
\end{enumerate}

\section{Proposed Method}\label{sec:Proposed_Method}

\footnotetext[1]{Figure icons obtained from flaticon.com}

\subsection{Approach}\label{subsec:Approach}
Unlike traditional recommendation scenarios~\cite{katlariwala2024product}, which center on product seeking, customer support conversations focus on problem solving.  
\name addresses this distinction with an implicit recommendation approach that analyzes conversation content instead of relying on user preferences.  
This enables \name to identify and recommend the most relevant SPCs (or specific products).

\subsection{Assumptions}\label{subsec:Assumptions}

\name leverages an existing online support platform (e.g., chatbot or forum) where users describe problems and request help.
It can be implemented as an add-on to these platforms, analyzing each support conversation and returning a prioritized list of SPC recommendations.

\subsection{Key Steps}\label{subsec:Key_Steps}

\name's input is a support conversation, i.e., a set of utterances between a user and a knowledgeable entity, either human or virtual.
As illustrated in \autoref{fig:methodology}, 
\name follows a three-step process in order to output a prioritized list of recommendations:

\subsubsection{Query Generation (Step 1)}\label{subsubsec:Query_generation} 

\begin{figure}[h]
  \centering
  \includegraphics[width=1.0\linewidth, trim={0 0 0 0},clip]{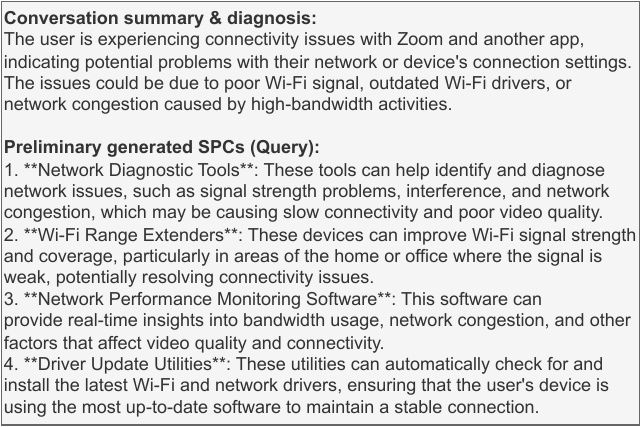}
  \vspace{-8pt}
  \setlength{\abovecaptionskip}{0pt}
  \setlength{\belowcaptionskip}{0pt}
  \caption{Query generation example from $DS^{CT}$.}
  \label{fig:query_generation_example}
\end{figure}

In this step (illustrated in the example in   \autoref{fig:query_generation_example}), given a conversation with multiple interactions, an LLM first generates a \emph{conversation summary and diagnosis} object, which concisely outlines the issue raised by the user, a root cause diagnosis, and plausible measures to take.
Then, based on the diagnosis, the LLM generates a \emph{query} object, which is a preliminary list of relevant SPCs that can help resolve the issue or prevent its reoccurrence.
Each SPC is also briefly explained, facilitating similarity search in the candidate retrieval step that follows. 

\subsubsection{Candidate Retrieval (Step 2)}\label{subsubsec:Candidate_generation}

\begin{figure}[!h]
  \centering
      \begin{tablenotes}
      \footnotesize
      \item Note: Bold entries indicate true-labeled SPCs.
    \end{tablenotes}
  \includegraphics[width=1.0\linewidth, trim={0 0 0 0},clip]{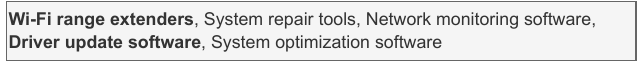}
  \vspace{-8pt}
  \setlength{\abovecaptionskip}{0pt}
  \setlength{\belowcaptionskip}{0pt}

  \caption{Candidate retrieval example from $DS^{CT}$.}
  \label{fig:candidates_retrieval_example}
\end{figure}

In this step (illustrated in the example in \autoref{fig:candidates_retrieval_example}), \name searches multiple designated databases (DBs) using the query generated in the previous step.
These DBs, equipped with an L2 index for efficient similarity search using an embedding model, differ from one another in terms of the aspect of the SPCs they emphasize or their source, thus enabling more diverse candidate retrieval than a single-index approach.
The final candidate set is formed by uniting the results from each index.

\subsubsection{Candidate Ranking (Step 3)}\label{subsubsec:Candidate-ranking} 
\begin{figure}[!h]
  \centering
      \begin{tablenotes}
      \footnotesize
      \item Note: Bold entries indicate true-labeled SPCs.
    \end{tablenotes}
  \includegraphics[width=1.0\linewidth, trim={0 0 0 0},clip]{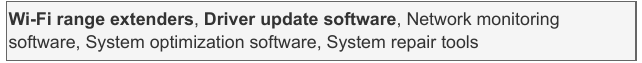}
  \vspace{-8pt}
  \setlength{\abovecaptionskip}{0pt}
  \setlength{\belowcaptionskip}{0pt}
  \caption{Candidate ranking example from $DS^{CT}$.}
  \label{fig:candidates_ranking_example}
\end{figure}

Given a set of retrieved candidates, the LLM ranks the SPCs by their ability to help resolve the diagnosis generated in the query generation step (see \autoref{fig:candidates_ranking_example}). 
To mitigate a possible position bias in ranking, \name employs a bootstrap approach~\cite{hou2024large}, concurrently repeating the ranking process three times with randomly shuffled orders of candidates.

\begin{table*}[t]
\caption{Datasets used to empirically evaluate \name.}
\label{tab:datasets}
\centering
\footnotesize
\resizebox{\textwidth}{!}{
{

\begin{tabular}{lp{1.89cm}p{2.77cm}p{0.7cm}p{0.9cm}p{7.6cm}p{1.64cm}}
\toprule
\textbf{Dataset} & \textbf{Domain} & \textbf{Source} & \textbf{Setting} & \textbf{\#Records} & \textbf{Example Question} & \textbf{Example SPC} \\
\midrule
$DS^{CT}$ & Cybersecurity Troubleshooting & Real conversations & Chatbot & 20 & "im in airport and i need to access my banking app do you think its safe to connect to available wifi" & VPN service \\
\midrule
$DS^{IS}$ & Information \newline Security & Stack Exchange network, IS community~\cite{InfoSecStack2025} & Online forum & 70 & "Precautions to secure (company) laptop \& mobile while traveling" & Mobile \newline antivirus \\
\midrule
$DS^{GE}$ & General Problem Solving & Synthetic & Chatbot & 224 & 
"... how to cope with my partner's snoring – it's been super tough lately. Any tips on how to handle this without causing much tension?"
& Sleep \newline headphones \\
\bottomrule
\end{tabular}%
}
}
\end{table*}

\section{Evaluation Method}\label{sec:Evaluation_Method}

\subsection{Creation of Datasets}\label{subsec:Dataset_Creation}

To evaluate \name, we constructed three datasets, which are summarized in \autoref{tab:datasets}:

\subsubsection{Cybersecurity Troubleshooting Dataset (\texorpdfstring{$DS^{CT}$}{DSCT})}\label{subsubsec:Cybersecurity_Troubleshooting}
PC users may occasionally encounter various technical issues, some of which stem from cybersecurity threats.
For example, PC slowness may be caused by software conflicts, outdated drivers, aging hardware, or cybersecurity attacks such as cryptojacking. 
To address this, we developed a cybersecurity-specialized chatbot in our lab and tasked students with troubleshooting a set of predefined complaints using only this chatbot. 
$DS^{CT}$ consists of these step-by-step cybersecurity troubleshooting conversations.

\subsubsection{Information Security Dataset (\texorpdfstring{$DS^{IS}$}{DSIS})}\label{subsubsec:Information_Security}
Stack Exchange~\cite{StackEx2025} is a network of nearly 200 question-and-answer (Q\&A) communities where millions collaborate monthly to ask questions, share knowledge, and solve problems across various technical and professional domains. 
Its most prominent community is the Stack Overflow community~\cite{StackOverflow2025}, which focuses on programming.
The Information Security (IS) community~\cite{InfoSecStack2025} on Stack Exchange, features discussions on cryptography, encryption, network security, and related topics. 
From this community, we extracted multiple Q\&A pairs in which the answer was accepted by the question's author.
Out of the 239 Q\&A pairs that we managed to annotate, we identified 70 pairs containing at least one product recommendation -- and these formed our final dataset.
In order to (1) prevent data leakage during testing and (2) ensure that the Candidate Generation step (Sec.~\ref{subsubsec:Candidate_generation}) relies solely on \name's conversation summary and diagnosis, we made sure to remove any mentions of specific product recommendations or specific solutions from the answers.

\subsubsection{General Problem-Solving Dataset (\texorpdfstring{$DS^{GE}$}{DSGE})}\label{subsubsec:General_Problem_solving} 
Following the LLM-based student-teacher interaction simulation framework proposed by Abbasiantaeb et al.~\cite{abbasiantaeb2024let}, we simulated support conversations with a chatbot specialized in a broad range of consultation topics, e.g., kitchen, pet care, and camping.
To simulate these conversations, we (1) generated synthetic users based on persona distributions, characterized by attributes such as age, gender, and occupation; and (2) instructed them to interact with an AI assistant (chatbot). 
Each conversation begins with the user describing a general problem encountered, followed by up to four Q\&A exchanges with the chatbot, enabling it to gather sufficient information.
The ground-truth SPC label is derived from the conversation generation prompt, which encapsulates both the essence of the problem and its root cause, to be communicated to the chatbot by the (synthetic) user. 
Although the user is aware of the cause, they are instructed not to disclose it explicitly. 
This setup helps maintain coherence and prevents divergence in the conversation.

To ensure conversation quality, we asked three experts to manually rate a random sample containing 10\% of $DS^{GE}$.
The rating was performed using the USR metrics for dialog generation~\cite{mehri2020usr} and achieved very high scores across all assessment dimensions.
The generated conversations received perfect scores for Understandable (range [0, 1], mean$\pm$standard deviation 1.00$\pm$0.00) and Maintains Context ([1, 3], 3.00$\pm$0.00), and high rating for Natural ([1, 3], 2.33$\pm$0.50) and Overall Quality ([1, 5], 4.01$\pm$0.66).
Gwet's AC2 coefficient ranged [0.84, 1.00], indicating near-perfect inter-rater agreement on these conversation quality metrics.

\subsection{Creation of Catalog DBs }\label{subsec:catalog_DBs_Creation}

\begin{figure}[t]
  \centering
  \includegraphics[width=1.0\linewidth, trim={0 0 0 10},clip]{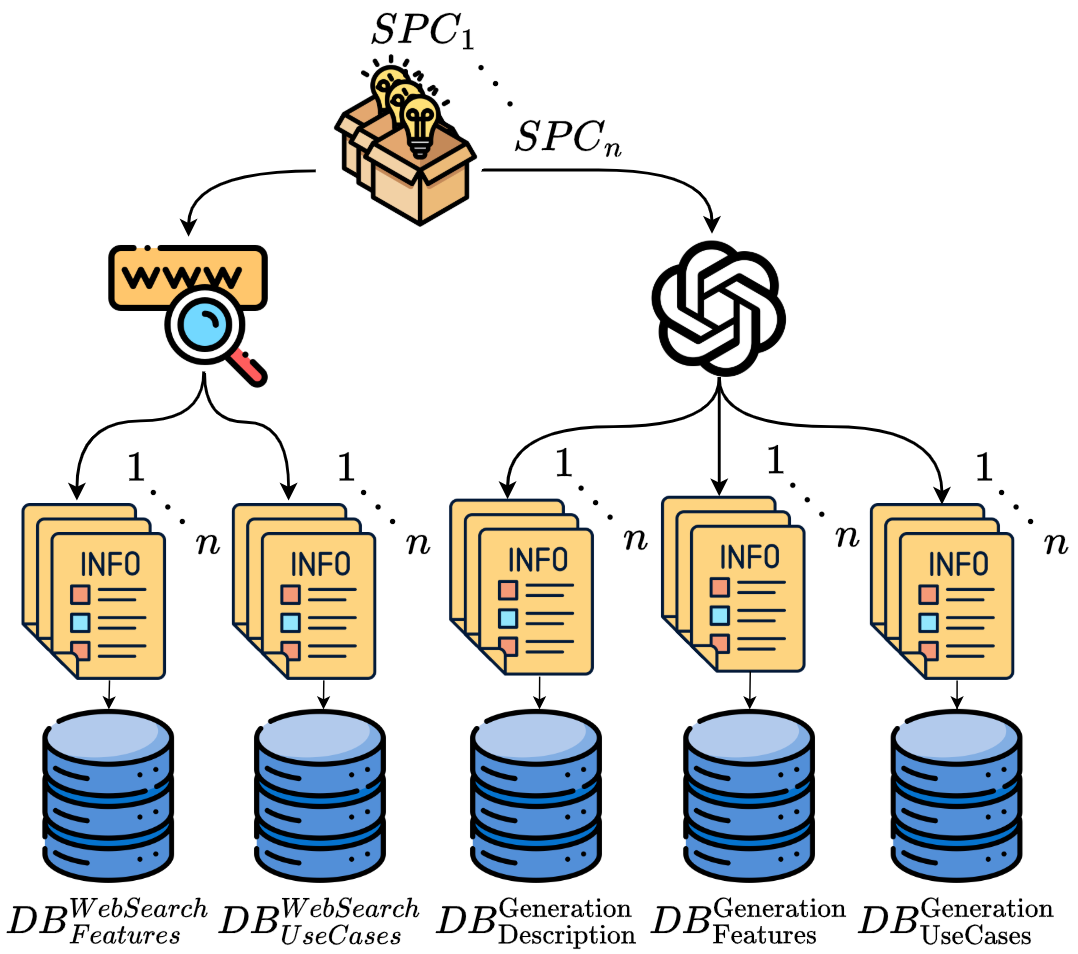}
  \caption{Catalog DB creation process\protect\footnotemark[1].}
  
  \label{fig:db_creation}
\end{figure}

As illustrated in \autoref{fig:db_creation}, we constructed five catalog DBs using both web search-based and LLM-based text generation methods. 
These approaches are complementary: search yields web pages deemed relevant by search engines, which likely contain useful information but may (1) include irrelevant content, or (2) lack comprehensive coverage by presenting only a few specific results.
In contrast, LLM-based text generation can synthesize information across multiple sources, offering broader summaries, which may come at the cost of factual inaccuracies.
In Sec.~\ref{subsec:Ablation_Study}, we present an ablation study to empirically assess the contribution of each catalog DB.

\subsubsection{Web Search-Based Catalog DBs Creation}\label{subsubsec:Retrieval_Based_Catalog_DB_Creation} 
Using Tavily~\cite{tavily}, a web search engine tailored for LLM retrieval-augmented generation (RAG) use cases, we constructed two catalog DBs: $DB^{WebSearch}_{Features}$ and $DB^{WebSearch}_{UseCases}$. For each SPC, Tavily was queried separately for each SPC's features and use cases, and each query returned five results.
The retrieved results for each SPC were then concatenated into two separate documents: one document containing all feature-related results, and the other containing all use case-related results.
The documents for each of the SPCs formed the respective DB.

\subsubsection{Generation-Based Catalog DBs Creation}\label{subsubsec:GEneration_Based_Catalog_DB_Creation} 
Using GPT-4o, we generated additional three catalog DBs. 
For each SPC, GPT-4o produced a brief description, a list of key features, and three example use cases. 
These outputs were stored in $DB^{\text{Generation}}_{\text{Descriptions}}$, $DB^{\text{Generation}}_{\text{Features}}$, and $DB^{\text{Generation}}_{\text{UseCases}}$, respectively.

\subsection{Experimental Setup}\label{subsec:Experimental_Setup}
We used GPT-4o mini to create $DS^{GE}$ -- both to generate users and the support conversations, where each user query and assistant response were generated independently.
We chose to use a high temperature (1.0) for these tasks to elicit diverse responses that better represent conversational scenarios. 
In contrast, for catalog DB creation and candidate search we used GPT-4o (and text-embedding-3-small~\cite{openai2023textembedding3small}, with a low temperature (0.3), to keep the outputs grounded and reduce variability in the content generated.
We implemented \name's pipeline on a laptop with an Intel i7-1365U 13th generation processor and 32 GB of RAM, using LangChain and LangSmith.
Access to GPT and Llama models was provided via Azure OpenAI Service and Amazon Bedrock, respectively.

\subsection{Performance Metrics}\label{subsec:Performance_Metrics}

We evaluated \name's performance using two metrics commonly used in recommender systems research~\cite{zhao2021recbole, zhao2024unlocking}: MRR@k, which measures how high a relevant SPC ranks among the first k suggestions; and Recall@k (abbreviated as R@k), which measures how many of the total relevant SPCs are successfully retrieved.
To align with both presentation strategies, we focused mainly on MRR@1 and R@3. We chose k=1 as most relevant for the "In-Chat" strategy for two reasons: (1) it may not be appropriate for a chatbot to recommend more than one SPC in a support conversation since recommendations are displayed directly within the conversation interface, and (2) most conversations contain one relevant SPC. 
Thus, MRR@1 effectively reflects how well \name retrieves and ranks this key recommendation. 
For "Related Items" representation strategy, which is applicable for multiple SPCs, and for a more comprehensive assessment, we also compute the MRR and recall at larger k values.

\section{Results}\label{sec:Results}

\subsection{Performance Across Datasets}\label{subsec:Performance_across_Datasets}

As can be seen in \autoref{tab:performance_comparison}, for each evaluated dataset both the MRR@k and R@k increase with k, although with diminishing effect.
The improvement in performance makes sense, since the likelihood of missing relevant SPCs decreases as more SPCs are retrieved.
The diminishing effect is a good indication that \name has already identified and ranked the most important recommendations, so as k increases, we see smaller improvements, if at all.

\begin{table}[t]
\caption{\name's performance across datasets.}
\label{tab:performance_comparison}
\footnotesize
\setlength{\tabcolsep}{4.5pt}
\begin{tabular}{lcccccc}
\toprule
\textbf{Dataset} & MRR@1 & MRR@3 & MRR@5 & R@1 & R@3 & R@5\\
\midrule
\shortstack[l]{$DS^{IS}$} & 0.82 & 0.87 & 0.87 & 0.68 & 0.83 & 0.87 \\[5pt]
\shortstack[l]{$DS^{CT}$} & 0.85 & 0.87 & 0.88 & 0.56 & 0.67 & 0.74 \\ [5pt]
\shortstack[l]{$DS^{GE}$} & 0.72 & 0.78 & 0.81 & 0.72 & 0.89 & 0.93 \\
\bottomrule
\end{tabular}
\end{table}

\subsection{Sensitivity to LLM and Embedding Models}\label{subsec:Sensitivity_to_LLM_and_Embedder}

\begin{figure*}[b]
  \centering
  \includegraphics[width=0.95\linewidth, trim={0 0 0 0},clip]{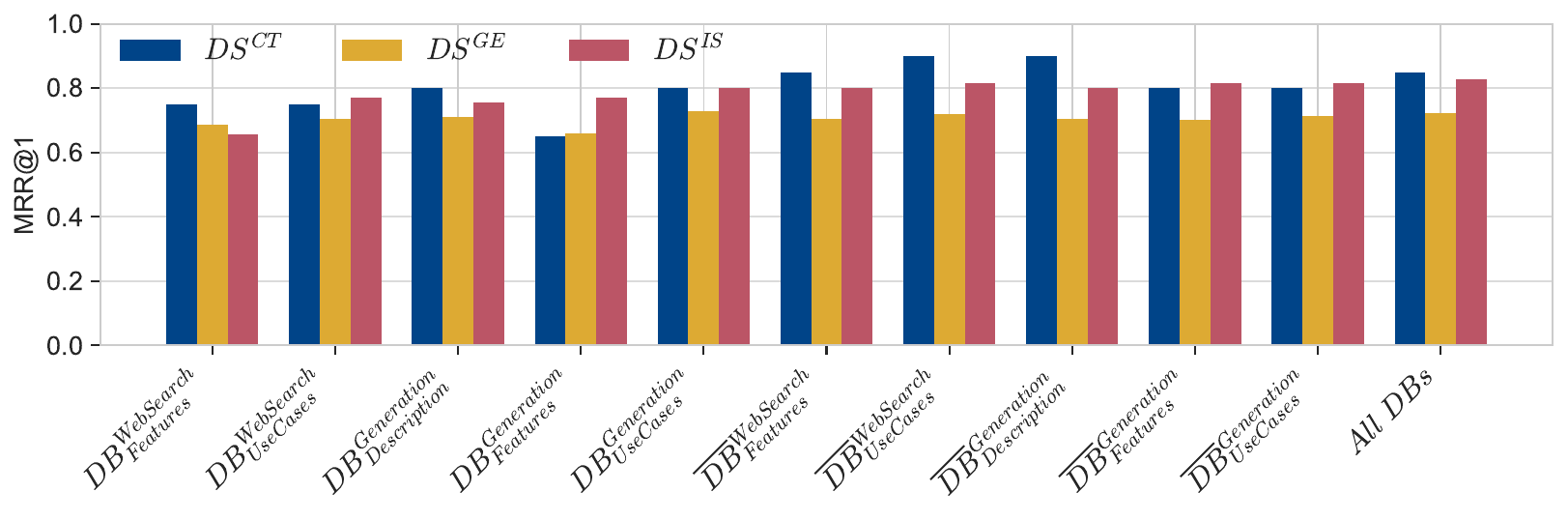}
  \setlength{\abovecaptionskip}{0pt}
  \setlength{\belowcaptionskip}{0pt}
  \renewcommand{\thefigure}{7}
  \caption{Ablation study results for various combinations of catalog DBs.}
  \label{fig:ablation}
\end{figure*}

The results presented in Sec.~\ref{subsec:Performance_across_Datasets} were obtained using a proprietary, state-of-the-art LLM and embedding model, respectively OpenAI's GPT-4o and text-embedding-3-small.
To evaluate \name's sensitivity to the underlying LLM and embedding model, we repeated the previous experiment with additional configurations.
That is, for the steps presented in Sec.~\ref{sec:Proposed_Method} we used two additional LLMs, GPT-4o mini (less costly) and Llama-3.3-70B-Instruct (open sourced), along with Multilingual-E5-Large-Instruct~\cite{wang2024multilingual} as an additional embedding model (open sourced, 560M parameters).
As can be seen in \autoref{tab:model_comparison}, GPT-4o outperformed GPT-4o mini and Llama-3.3-70B-Instruct in almost all cases (frequently with a high margin), as did text-embedding-3-small compared to Multilingual-E5-Large-Instruct (though with a smaller margin).
Moreover, the combination of GPT-4o with text-embedding-3-small yielded the highest MRR@1 for 3 out of 3 datasets and the highest R@3 for 2 out of 3 datasets.
The superiority of GPT-4o in this experiment is consistent with prior research~\cite{sanghera2025high}, including  recommender systems research~\cite{vajjala2024cross}.

\subsection{Sensitivity to Conversation Length}\label{subsec:Sensitivity_to_Conversation_Length}

\begin{table}[t]
\footnotesize
\caption{Performance across LLMs and embedding models.}
\label{tab:model_comparison}
\begin{threeparttable}
\setlength{\tabcolsep}{2.3pt} 
\begin{tabular}{llcccccc}
\toprule
\multirow{2}{*}{\textbf{LLM}} & \multirow{2}{*}{\textbf{\shortstack[l]{Embed.\\model}}} & \multicolumn{2}{c}{$DS^{IS}$} & \multicolumn{2}{c}{$DS^{CT}$} & \multicolumn{2}{c}{$DS^{GE}$} \\
\cmidrule(lr){3-4} \cmidrule(lr){5-6} \cmidrule(lr){7-8}
& & MRR@1 & R@3 & MRR@1 & R@3 & MRR@1 & R@3 \\
\midrule
\multirow{2}{*}{4o} & multi-e5  & 0.74 & \textbf{0.83} & 0.60 & 0.56 & 0.70 & \textbf{0.90} \\
& embed-3 & \textbf{0.82} & \textbf{0.83} & \textbf{0.85} & 0.67 & \textbf{0.72} & 0.89 \\
\midrule
\multirow{2}{*}{4o mini} & multi-e5  & 0.51 & 0.60 & 0.55 & 0.55 & 0.63 & 0.85 \\
& embed-3 & 0.55 & 0.67 & 0.65 & 0.64 & 0.61 & 0.84 \\
\midrule
\multirow{2}{*}{Llama} & multi-e5  & 0.62 & 0.66 & 0.55 & 0.56 & 0.66 & 0.85 \\
& embed-3 & 0.62 & 0.73 & 0.80 & \textbf{0.69} & 0.67 & 0.84 \\
\bottomrule
\end{tabular}
\begin{tablenotes}
\item Note: Bold values indicate the best configuration for each metric.
\end{tablenotes}
\end{threeparttable}
\end{table}

\begin{figure}[t]
  \centering
  \includegraphics[width=0.89\linewidth, trim={0 0 0 0},clip]{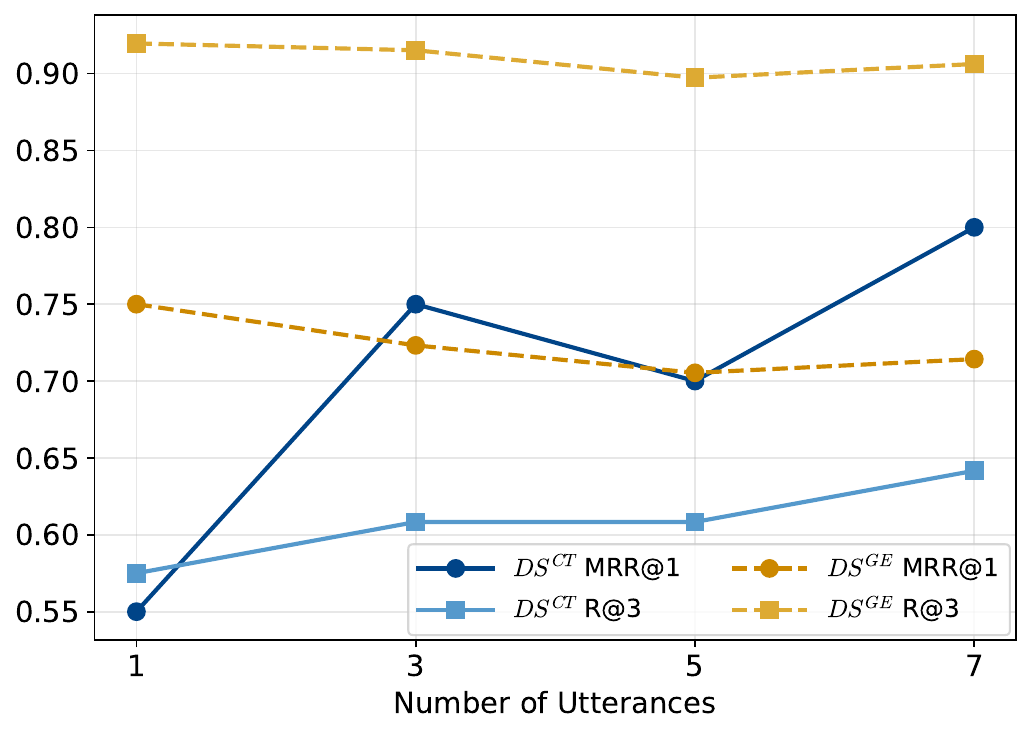}
\setlength{\abovecaptionskip}{0pt}
  \setlength{\belowcaptionskip}{0pt}
  \vspace{3.5pt}
  \renewcommand{\thefigure}{6}
  \caption{Performance as a function of conversation length.}
  \label{fig:sensitivity}
\end{figure}

The experimental results discussed thus far were achieved using all available utterances in each conversation.
To assess \name's sensitivity to conversation length, \autoref{fig:sensitivity} shows the MRR@1 and R@3 scores as functions of the number of utterances in the conversation.
Since $DS^{IS}$ contains only one user message followed by an assistant response, without further utterances, it is not included in this analysis, which only pertains to $DS^{CT}$ and $DS^{GE}$.
The experimental results show that on $DS^{CT}$, both the MRR@1 and R@3 increase with the number of utterances.
This makes sense, as cybersecurity troubleshooting often requires
iteratively ruling out possible root causes.
In other words, the less irrelevant root causes considered by the chatbot, the fewer irrelevant SPCs recommended.
In comparison, the user complaints and questions in $DS^{GE}$ were simpler (e.g., planning holiday meals), with most of the information provided in the first question, so high MRR@1 and R@3 values were quickly achieved.

\subsection{Ablation Study}\label{subsec:Ablation_Study}
Our literature review did not reveal any prior research (with or without an experimental dataset made public) on conversational recommendations based on implicit user needs; in fact, we found no research whatsoever on conversational recommendations that are not based on user preferences.
Hence, since we were unable to directly compare \name to a public benchmark, and in order to better explore \name's potential while assessing the relative importance of its components, we performed an ablation study, which is described below.

\begin{figure}[t]
  \centering
  \includegraphics[width=0.9\linewidth, trim={0 0 0 0},clip]{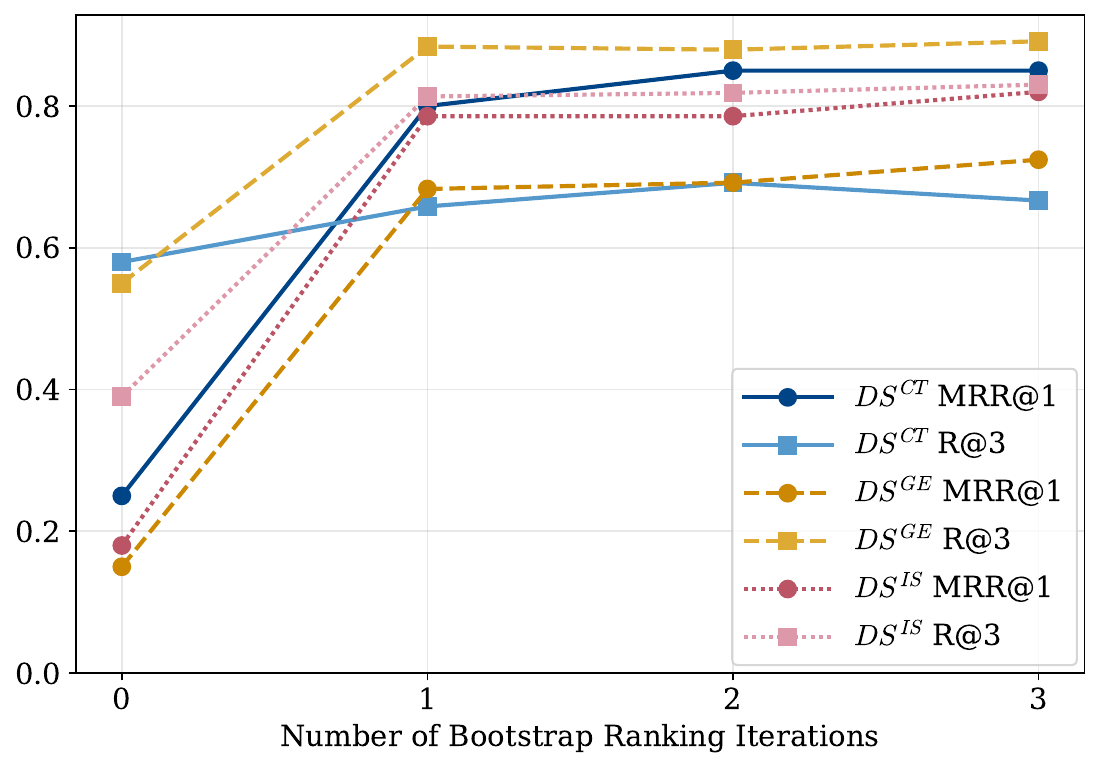}
  \vspace{-8pt}
  \caption{Performance as a function of bootstrap iterations.}
  \label{fig:bootstrap_ablation}
\end{figure}

\begin{figure*}[!b]
  \centering
  \begin{subfigure}[b]{0.33\linewidth}
    \centering
    \includegraphics[width=\linewidth, trim={0 0 0 0},clip]{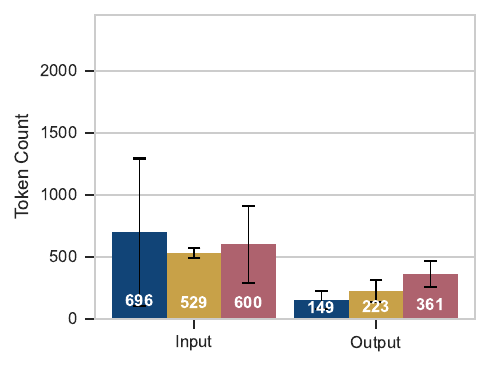}
    \caption{Summary \& diagnosis (Step 1)}
    \label{fig:summary_diagnosis_time}
  \end{subfigure}
  \hfill
  \begin{subfigure}[b]{0.33\linewidth}
    \centering
    \includegraphics[width=\linewidth, trim={0 0 0 0},clip]{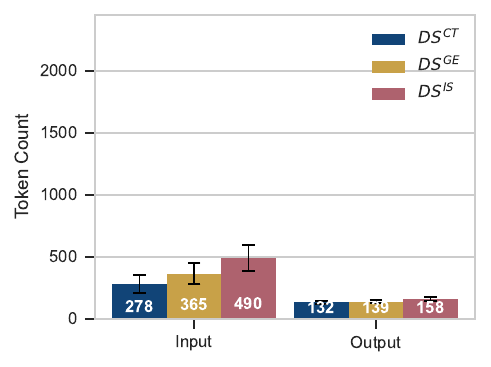}
    \caption{Preliminary generated SPCs (Step 1)}
    \label{fig:initial_spcs}
  \end{subfigure}
  \hfill
  \begin{subfigure}[b]{0.33\linewidth}
    \centering
    \includegraphics[width=\linewidth, trim={0 0 0 0},clip]{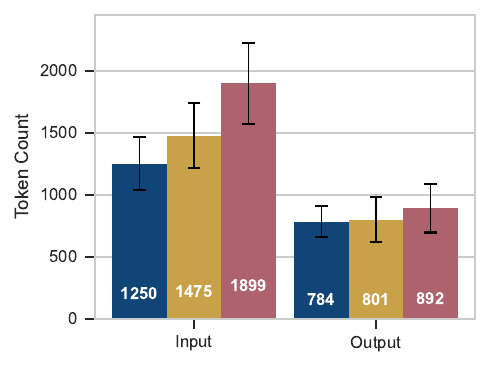}
    \caption{Bootstrap ranking (Step 3)}
    \label{fig:ranking_time}
  \end{subfigure}
  
  \renewcommand{\thefigure}{10}
  \caption{Token consumption overhead analysis.}  \label{fig:token_overhead}
\end{figure*}

\begin{figure}[t]
  \centering
  \includegraphics[width=0.95\linewidth, trim={0 0 0 0},clip]{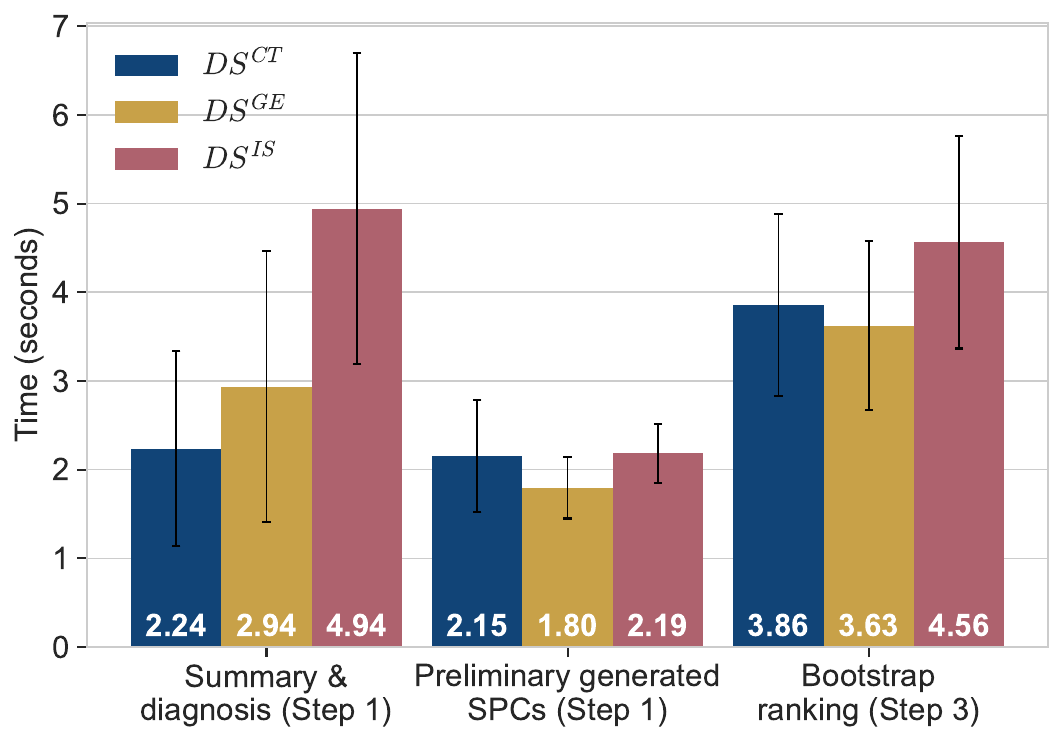}
  \setlength{\abovecaptionskip}{0pt}
  \setlength{\belowcaptionskip}{0pt}
  \renewcommand{\thefigure}{9}
  \caption{Time overhead analysis.}
  \label{fig:time_overhead}
\end{figure}

\subsubsection{Importance of the Various Catalog DBs}\label{subsubsec:Importance_of_the_Various_Catalog_DBs}

As described in Sec.
\ref{subsec:catalog_DBs_Creation}, the catalog DBs contain SPC features, descriptions, and use cases, obtained using web search and LLM-based text generation.
In this experiment, we compared the performance across all three evaluated datasets when using all five DBs together (denoted as \textit{$All\ DBs$}) against using each DB individually, as well as against using every combination of four DBs.
For example, $DB^{WebSearch}_{Features}$ denotes including \emph{only} this DB, while $\overline{DB}^{WebSearch}_{Features}$ denotes including all other (four) DBs \emph{except for} this DB.
As can be seen in \autoref{fig:ablation}, on $DS^{IS}$ the $All\ DBs$ configuration outperforms all other configurations, similarly to $DS^{GE}$, and on $DS^{CT}$, the $All\ DBs$ configuration works best in 9 out of 11 of tested configurations; based on this, we can conclude that each of the five DBs contributes to \name's performance.
In future implementations, if overhead limitations, such as token consumption costs, are more restrictive, the best option would be to use any of the Use Cases DBs (based on web search or LLM generation), as retrieving SPC candidates based on their typical use cases is more effective than retrieval based on the SPCs' features or descriptions.

\subsubsection{Importance of Candidates' Bootstrap Ranking}\label{subsubsec:Importance_of_Candidates_Bootstrap_Ranking}

As discussed in Sec.~\ref{subsec:overhead}, the third step of \name -- candidate ranking -- introduces considerable overhead in both cost and latency.
To assess the necessity of this step, we evaluated \name with zero to three bootstrap 
ranking iterations. 
The results, shown in \autoref{fig:bootstrap_ablation}, reveal a marked performance drop across all datasets when the candidate ranking is disabled, i.e., when \name relies solely on the candidate SPCs retrieved from the catalog DBs, ranked by their similarity to the generated query (Sec.~\ref{subsubsec:Query_generation}). 
Unlike Hou et al.~\cite{hou2024large}, who advocate for at least three ranking iterations, our findings indicate that even one or two iterations substantially improve performance, though still falling short of the optimal results achieved with three.

\subsection{Overhead}\label{subsec:overhead}

Figures~\ref{fig:time_overhead} and~\ref{fig:token_overhead} present \name's overhead in terms of time (seconds) and token consumption, respectively indicating computation and monetary expenses. Although the overhead was typically higher for $DS^{IS}$ than it was for $DS^{CT}$ and $DS^{GE}$, in most cases the differences were relatively small.

While bootstrap ranking the retrieved SPC candidates is the most resource-consuming step, this step also greatly affects \name's performance (as discussed in Sec.~\ref{subsubsec:Importance_of_Candidates_Bootstrap_Ranking}).
The overhead of that step can be reduced in several ways, including the use of a local LLM, which may result in shorter and more consistent waiting times.

\section{Related Work}\label{sec:Related_Work}

\begin{table*}
\caption{Comparison of reviewed conversational and LLM-based recommender systems.}
\label{tab:related_work_comparison}
\footnotesize
\resizebox{\textwidth}{!}{ 
\begin{tabular}{p{0.65cm}p{1.08cm}p{3.34cm}p{1.99cm}p{3.87cm}p{2.97cm}p{2.71cm}}
\toprule
\textbf{Paper} & 
\textbf{Name} & \textbf{Recommender Type} & \textbf{
Paradigm} & \textbf{Application Domains} & \textbf{Rec. Intent Assumption?} & \textbf{Conversation Control?} \\
\midrule
\cite{wang2022towards} & UniCRS & Conversational & Preference-based & Movies & Yes & Yes \\ [2pt]
\cite{deng2023unified} & MG-CRS & Conversational & Preference-based & Movies, music, restaurants & No & Yes \\ [2pt]
\cite{wang2021recindial} & RecInDial & Conversational & Preference-based & Movies & Yes & Yes \\ [2pt]
\cite{xu2021adapting} & FPAN & Conversational & Preference-based & Music, Yelp dataset
& Yes & Yes \\ [2pt]
\cite{wang2024recmind} & RecMind & LLM-powered agent & Preference-based & E-commerce, Yelp dataset & Yes & N.A. \\ [2pt]
\cite{wang2024macrec} & MACRec & Multi-agent collaborative & Preference-based & E-commerce, movies, social media & Yes & N.A. \\ [2pt]
\cite{shu2024rah} & RAH! & Human-centered, LLM agents & Personality-based & Movies, books, games & Yes & N.A. \\ [2pt]
This & \name & Support conversation add-on & Need-based & Support conversations & No & No \\
\bottomrule
\end{tabular}
}
\end{table*}
 
\subsection{Chatbots in Customer Service and Support}\label{subsec:Chatbots_in_Customer_Service_and_Support}

Live chat interfaces have become a widely adopted channel for delivering real-time customer service~\cite{shang2024personalized}. 
Customers increasingly rely on these platforms to access information or receive assistance, and the immediacy of live chat strongly influences customer trust and satisfaction~\cite{adam2021ai}. 
Hardalov et al.~\cite{hardalov2018towards} examined the automation of customer support using conversational agents, comparing methods such as information retrieval, sequence-to-sequence (Seq2Seq) models, attention mechanisms, and the transformer architecture. 
Other studies have investigated domain-specific applications in sectors such as banking~\cite{weerabahu2019digital} and healthcare~\cite{ayanouz2020smart}, employing various deep learning and natural language processing techniques. 
More recently, LLMs have been applied to customer support tasks, including fine-tuning OpenAI's GPT-4 for e-commerce~\cite{santosh2024leveraging} and deploying Google's Flan-T5 XXL model for real-time assistance~\cite{pandya2023automating}. 
Additional research has explored LLM-driven approaches to create context-aware and personalized support chatbots~\cite{santosh2024leveraging}. 
While these methods aim to enhance the quality of customer service, they typically do not utilize conversation data to recommend relevant products or services -- an approach that could further support users and drive business growth at the same time.

\subsection{Conversational Recommender Systems}\label{subsec:Conversational_Recommender_System}
CRSs collect user preferences and deliver personalized suggestions through interactive natural language dialogues~\cite{sun2018conversational}.
CRSs are broadly categorized into attribute-based and open-ended systems.
Attribute-based CRSs refine user preferences through explicit, attribute-driven Q\&A interactions, aiming to identify the optimal item(s) in minimal rounds~\cite{lei2020interactive, ren2021learning, xu2021adapting}.
In contrast, open-ended CRSs such as RecInDial~\cite{wang2021recindial} and UniCRS~\cite{wang2022towards} support free-form conversations, enabling more flexible user interactions.
To enhance performance, UniCRS incorporates dialogue history as contextual input and knowledge graph entities as external information to jointly address recommendation and dialogue tasks.
Deng et al.~\cite{deng2023unified} proposed UniMIND, a Seq2Seq model that unifies multiple CRS goals -- including chitchat, question answering, topic prediction, and recommendation—via prompt-based learning.
However, UniMIND's multi-goal approach can detract from recommendation progress, especially when chitchat dominates.
Fundamentally, CRSs drive recommendations by steering conversations to elicit user preferences and background, assuming that users engage with the intent of receiving suggestions.
In contrast, \name assumes no underlying purchasing intent and exerts no control over the conversation’s direction. 
Functioning as an add-on, \name analyzes the support conversation, and recommends the most suitable SPCs -- the ones capable of addressing the identified issue.

\subsection{LLM-Based Recommender Systems}\label{subsec:LLM_Based_Recommender_Systems}

LLMs have been integrated in various components of recommender systems, including feature engineering, user and item embeddings, scoring, ranking, or even functioning as agents that guide the recommendation process itself~\cite{lin2025can}. 
For item embedding, TedRec~\cite{xu2024sequence} performs sequence-level semantic fusion of textual and ID features for sequential recommendation, while NoteLLM~\cite{zhang2024notellm} combines note semantics with collaborative signals to produce note embeddings. 
In contrast, Chen et al.~\cite{chen2024hllm} proposed a hierarchical approach where an LLM extracts features from item descriptions and converts them into compact embeddings, reducing computational overhead.
Other studies leveraged autonomous, LLM-powered agents. 
Unlike methods that merely prompt LLMs with user history, these approaches exploit advanced agentic capabilities such as planning and tool use. 
RecMind~\cite{wang2024recmind} performs multi-path reasoning through LLM planning, MACRec~\cite{wang2024macrec} features collaboration among specialized agents, and RAH~\cite{shu2024rah} is a human-centered framework in which multiple LLM-based agents mediate between users and recommender systems to align suggestions with a user's personality and reduce cognitive load.

While most LLM-based systems focus on predicting what users might \emph{like} based on past behavior or preferences (see \autoref{tab:related_work_comparison}), we propose a paradigm shift: understanding what the user \emph{needs}. 
\name identifies recommendation opportunities by detecting needs -- explicitly expressed, implicitly inferred, or derived from interactions between a user and  a knowledgeable entity, either human or virtual.

\section{Discussion}\label{sec:Discussion}

\subsection{Key Insights}\label{subsec:Key_Insights}
\name's results are promising, with  a mean of 0.8 for both MRR@1 and R@3 achieved using a state-of-the-art LLM and embedding model. 
These performance levels are typically reached early in a conversation, highlighting \name's ability to produce accurate recommendations quickly, after just a short interaction with a customer/user.
Our ablation study confirms that each of \name's components contributes to its overall performance, although  some overhead is incurred (Sec.~\ref{subsec:overhead}). 
To mitigate time overhead, we recommend using more powerful hardware, and using parallel LLM calls for bootstrap ranking.
To reduce token consumption, our findings indicate
that even one or two iterations of bootstrap ranking substantially improve performance,
though still falling short of the optimal results achieved with three.

\subsection{Research Limitations and Future Work}\label{subsec:Research_Limitations_and_Future_Work}
While the results of this study are promising, their generalizability might be affected by factors such as the modest size of the datasets and the limited domain diversity.
In future work we will evaluate \name on additional, larger conversational datasets drawn from a variety of support scenarios.
We have also identified three other interesting research directions: (1) conducting an online experiment to examine user conversion rates after receiving recommendations from \name; (2) optimizing the \emph{timing} of the recommendation within the support conversation; and (3) refining the chatbot's phrasing of recommendations within the "In-Chat" presentation strategy.
In addition, given \name's sensitivity to the underlying LLM selection (Sec.~\ref{subsec:Sensitivity_to_LLM_and_Embedder}), future research could explore the use of 
newly released LLMs, particularly open-source models, that can run locally to both reduce costs and enhance privacy.

\section{Conclusion}\label{sec:Conclusion}
In this paper, we introduced \name, a novel LLM-based recommender system that, unlike traditional preference-based approaches, implicitly infers user needs through conversation.
It then suggests relevant SPCs (or specific products) potentially delivering value to both the customer and the business operating the support chatbot.
Our comprehensive quantitative evaluation performed on datasets from three application domains demonstrated \name's effectiveness, analyzed its sensitivity to various parameters, and assessed the relative contributions of its components.
In future work, we plan to broaden our experimental scope and collect online behavioral data, so that we can optimize the timing and refine phrasing of recommendations for the "In-Chat" presentation strategy to maximize conversion rates.

\bibliographystyle{ACM-Reference-Format}
\bibliography{sample-base}

\end{document}